\documentclass[conference]{IEEEtran}
\IEEEoverridecommandlockouts
\usepackage{cite}
\usepackage{amsmath,amssymb,amsfonts}
\usepackage{algorithmic}
\usepackage{graphicx}
\usepackage{textcomp}
\usepackage{xcolor}
\usepackage{multirow}
\usepackage{booktabs}
\usepackage{pifont}
\usepackage[normalem]{ulem}
\usepackage{url}
\usepackage{caption}
\usepackage[edges]{forest}
\usepackage{tikz}
\usetikzlibrary{positioning, shapes, arrows.meta}

\def\BibTeX{{\rm B\kern-.05em{\sc i\kern-.025em b}\kern-.08em
    T\kern-.1667em\lower.7ex\hbox{E}\kern-.125emX}}
\begin{document}

\title{Robustness Requirement Coverage using a Situation Coverage Approach for Vision-based AI Systems}

\author{\IEEEauthorblockN{1\textsuperscript{st} Sepeedeh Shahbeigi}
\IEEEauthorblockA{\textit{Department of Computer Science} \\
\textit{University of York}\\
York, United Kingdom \\
Sepeedeh.shahbeigi@york.ac.uk}
\and
\IEEEauthorblockN{2\textsuperscript{nd} Nawshin Mannan Proma}
\IEEEauthorblockA{\textit{Department of Computer Science} \\
\textit{University of York}\\
York, United Kingdom \\
nawshinmannan.proma@york.ac.uk}
\and
\IEEEauthorblockN{3\textsuperscript{rd} Victoria Hodge}
\IEEEauthorblockA{\textit{Department of Computer Science} \\
\textit{University of York}\\
York, United Kingdom \\
victoria.hodge@york.ac.uk}
\and
\IEEEauthorblockN{4\textsuperscript{th} Richard Hawkins}
\IEEEauthorblockA{\textit{Department of Computer Science} \\
\textit{University of York}\\
York, United Kingdom \\
richard.hawkins@york.ac.uk}
\and
\IEEEauthorblockN{5\textsuperscript{th} Boda Li}
\IEEEauthorblockA{\textit{WMG} \\
\textit{University of Warwick}\\
Coventry, United Kingdom \\
Boda.Li.4@warwick.ac.uk}
\and
\IEEEauthorblockN{6\textsuperscript{th} Valentina Donzella}
\IEEEauthorblockA{\textit{School of Engineering and Materials Science} \\
\textit{Queen Mary University of London}\\
London, United Kingdom \\
v.donzella@qmul.ac.uk}
}

\maketitle

\begin{abstract}
AI-based robots and vehicles are expected to operate safely in complex and dynamic environments, even in the presence of component degradation. In such systems, perception relies on sensors such as cameras to capture environmental data, which is then processed by AI models to support decision-making. However, degradation in sensor performance directly impacts input data quality and can impair AI inference. Specifying safety requirements for all possible sensor degradation scenarios leads to unmanageable complexity and inevitable gaps. In this position paper, we present a novel framework that integrates camera noise factor identification with situation coverage analysis to systematically elicit robustness-related safety requirements for AI-based perception systems. We focus specifically on camera degradation in the automotive domain. Building on an existing framework for identifying degradation modes, we propose involving domain, sensor, and safety experts, and incorporating Operational Design Domain specifications to extend the degradation model by incorporating noise factors relevant to AI performance. Situation coverage analysis is then applied to identify representative operational contexts. This work marks an initial step toward integrating noise factor analysis and situational coverage to support principled formulation and completeness assessment of robustness requirements for camera-based AI perception.
\end{abstract}


\section{Introduction} % 1.5 column
AI components are increasingly used in safety-critical systems such as autonomous vehicles and robots, which must operate reliably under diverse and potentially degraded conditions. However, AI-based perception, particularly when reliant on vision sensors, is highly sensitive to such variations~\cite{chan2020framework}. This makes their safety assurance particularly challenging.

Traditional safety engineering mitigates risk by identifying hazards and deriving explicit safety requirements. In AI systems, especially those involving perception, this approach becomes intractable due to the open-ended nature of the input space and their data-driven performance under noise and environmental stressors~\cite{molloy2024hazard}.

To address this, we propose a structured method that constrains the infinite input space and systematically defines variations in the operational context (i.e., the set of environmental and system conditions under which the AI component operates). These contextual variations constitute the basis for robustness requirements on the AI component. Building on prior work that classifies camera noise factors using a P-Diagram-based framework~\cite{li2022analysis}, we extend this foundation to support systematic robustness derivation. We adopt the situation coverage framework~\cite{proma2025scaloft}, not to generate test cases, but to formally represent the space of degraded operating conditions and guide robustness requirements elicitation.

Our motivation is twofold: (1) situation coverage provides a principled and extensible way to model sensor degradation factors and their combinations; and (2) this structured representation enables systematic identification of the performance boundaries within which AI-based perception must remain safe. To our knowledge, this is the first approach to integrate noise factor identification with situation coverage to derive robustness safety requirements in perception under degradation.

While existing frameworks~\cite{hawkins2021guidance, sotif2022, burton2017making} recognise the importance of robustness, they do not specify how to identify the full range of degraded conditions in a tractable, systematic way. Situation coverage addresses this gap by enabling comprehensive consideration of environmental variations and their interactions. Although previously applied to system-level testing~\cite{nawshin2024}, it has not yet been used for structured robustness requirement elicitation in AI-based perception systems.

\section{Automotive Camera Noise Factors Identification} %1 column

\begin{figure}[ht]
    \centering
    \includegraphics[width=1\linewidth]{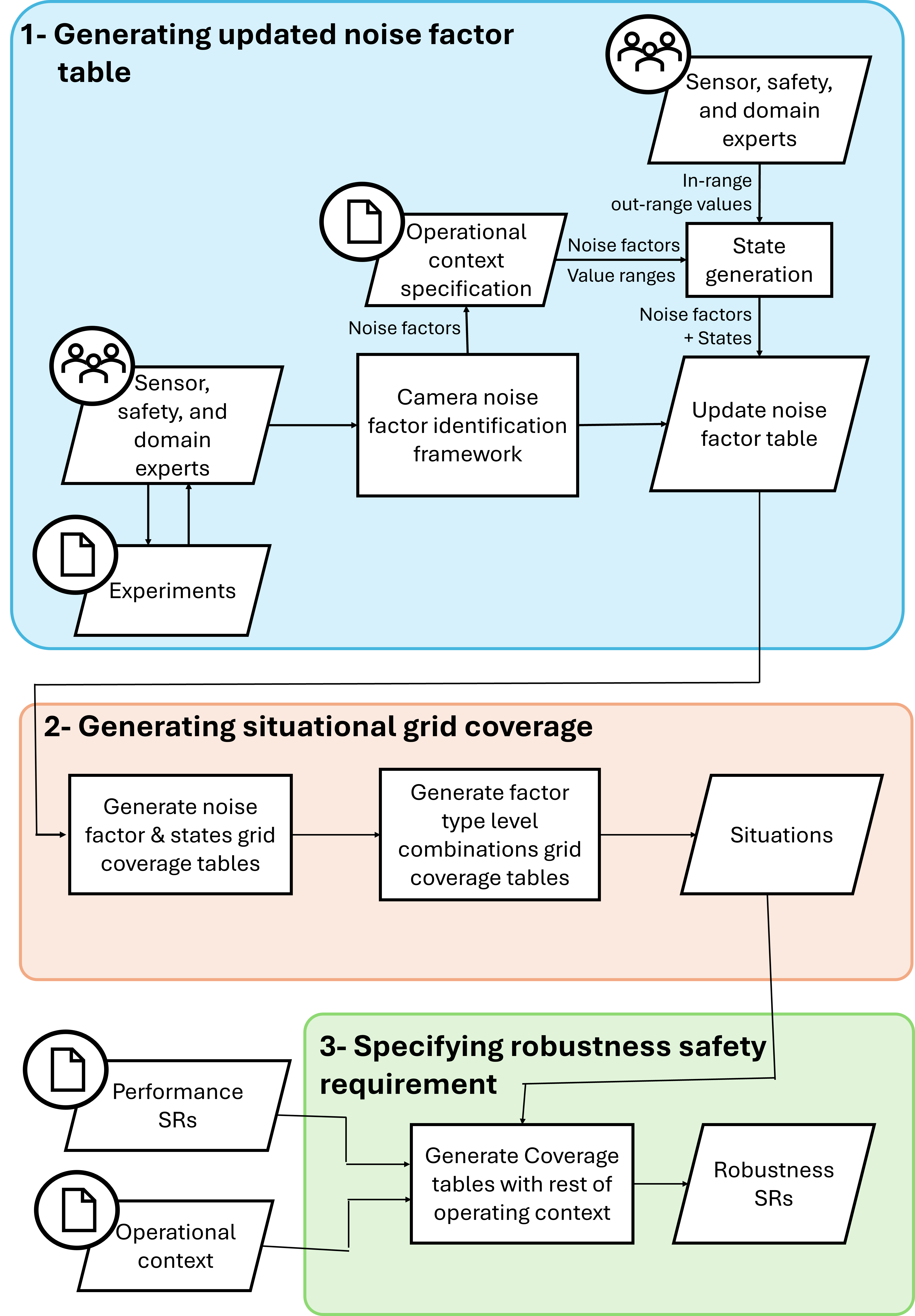}
    \caption{Framework overview for noise-based requirement derivation.}
    \label{fig:framework}
\end{figure}

Camera-based perception systems are highly sensitive to both internal and external sources of variation, which can degrade image quality and impair the performance of AI models. To support systematic identification and analysis of such effects, Li et al.~\cite{li2022analysis} proposed a Parameter Diagram (P-Diagram) approach to classify noise factors affecting automotive camera sensors. We adopt this method within our framework to derive traceable and structured safety requirements for perception under degradation.

The list of camera noise factors~\cite{li2022analysis} was developed through practical experience and expert consultation. It captures a broad range of systematic or recurring sources of sensor degradation. Currently, it does not include corner cases such as surface reflections that result in object ghosting, which fall outside the scope of regular degradation patterns.

The P-Diagram approach categorises noise factors into five groups. The first category, \textbf{piece-to-piece} variation, refers to manufacturing or assembly inconsistencies, such as sensor–lens misalignment or variability in lens material properties. The second, \textbf{change over time}, includes degradation mechanisms like electronic ageing and mechanical loosening due to prolonged vibration. The third, \textbf{usage}, captures operational stressors such as scratches, sensor misplacement, or the accumulation of contaminants during vehicle operation or maintenance. The fourth, \textbf{environmental factors}, covers external conditions that can lead to camera sensor degradation. Finally, \textbf{system interactions} refer to integration-related effects, such as electromagnetic interference, unstable power supply, or optical distortion introduced by the vehicle’s windshield curvature. Each identified factor is evaluated for its impact on key sensor outputs: frame rate (FR), pixel intensity ($I_{RGB}$), pixel position ($P_{(x,y)}$), and dropped frames (DF).

\section{The Grid Coverage-based Framework} % 1 page
Our framework, illustrated in Fig.~\ref{fig:framework}, comprises three high-level stages. Stage~1 constructs a noise factor table that includes relevant degradation factors, their discrete operational states, and inter-factor dependencies. This table focuses on factors that are relevant to the AI component within the specific application and task. For each factor, discrete states are defined based on the operating context specifications. Operating context refers to the environmental, spatial, and temporal conditions under which the system must operate safely.

Stage~2 generates situation coverage grids, enumerating all possible situations resulting from combinations of the noise factors defined in Stage~1. In this work, we adopt the term situation to refer specifically to aspects of the operating context related to camera degradation.

Stage~3 uses the situational grid table from Stage~2 to derive coverage tables that span the full set of operational context attributes. Each row in these coverage tables corresponds to a robustness safety requirement for the AI perception system.

\begin{table*}[ht]
\centering
\renewcommand{\arraystretch}{1.8} % Adjust this value as needed
\caption{Noise factor states and dependencies}
\label{tab:noisefactor}
\begin{tabular}{|l|l|l|l|}
\hline
\textbf{Factor Type} & \textbf{ID / Noise Factor} & \textbf{State(s)} & \textbf{Dependency} \\
\hline
\multirow{5}{*}{Piece to Piece}
 & 1. Alignment & False (Normal), True (Misalignment) & — \\
 & 2. Fabrication Variability & False (Normal), Low, High & — \\
 & 3. Lens Shape, Purity & False (Normal), Low, High & — \\
 & 4. Dark Current Variability & False (Normal), True (High) & — \\
 & 5. Image Signal Processing (ISP) & False (Normal), True & — \\
\hline
\multirow{4}{*}{Change over Time}
 & 1. Ageing of Electronics & False (Normal), True (High) & — \\
 & 2. Degradation of Lens & False (Normal), Low, High & — \\
 & 3. Vibration of Mounting & False (Normal), True & Co-occurrence with Misplacement \\
 & 4. Pollutant Ingress & False (Normal), Low, High & — \\
\hline
\multirow{6}{*}{Usage}
 & 1. Misplacement of Sensor & False (Normal), True & — \\
 & 2. Vehicle Impact & False (Normal), True & Co-Occurrence with Misplacement \\
 & 3. Chemicals / Contaminants & False (Normal), True & — \\
 & 4. Obstructions & False (Normal), Low, High & — \\
 & 5. Lens Scratch & False (Normal), Low, High & Co-occurrence with Obstructions \\
 & 6. Vehicle Dynamic Settings & False (Normal), True & Co-occurrence with Misplacement \\
\hline
\multirow{3}{*}{Environment}
 & 1. Sensor Saturation / Depletion & False (Normal), High & — \\
 & 2. Extreme Temperature & False (Normal), Low, High & — \\
 & 3. Low Illumination & False (Normal), High & — \\
\hline
\multirow{4}{*}{System Interaction}
 & 1. Malicious Attacks & False (Normal), True & — \\
 & 2. Windshield Distortion & False (Normal), Low, High & — \\
 & 3. Power Supply & False (Normal), True & — \\
 & 4. Electromagnetic Interference (EMI) & False (Normal), True & — \\
\hline
\end{tabular}
\end{table*}

\subsection{Stage~1: Generating an Updated Noise Factor Table}
Stage~1 focuses on identifying and organising the noise factors that affect camera sensors. We build on the taxonomy of thirty noise factors proposed by Li et al.~\cite{li2022analysis}, extending it to better suit safety-critical AI-based perception tasks. Our process begins with expert consultation to assess the relevance and severity of various degradation sources, taking into account the specific operational context and perception functions. As this is a proposed framework, expert input is used illustratively to highlight how this process would be conducted in practice. For instance, we assume consultation with sensor engineers and safety assessors to validate factor relevance. For assurance purposes, merely listing degradation sources is insufficient; it is essential to understand how each factor impacts the camera image data, as this directly affects AI performance. Therefore, this stage prioritises the analysis of image-level effects to identify degradation sources likely to impair perception.

To support this, we classify each noise factor based on its effect on image-level outputs, specifically whether it alters pixel intensity (\(I_\text{RGB}\)), pixel position (\(P_{(x,y)}\)), or both, following the methodology in~\cite{li2022analysis}. Factors that only influence FR or DF are excluded, as they do not directly affect the spatial or visual content processed by the AI. In some cases, the effect of a factor may not be known \emph{a priori} and may require targeted testing or simulation as illustrated by \emph{experiments} block in Fig.~\ref{fig:framework}.

We also capture dependencies between degradation factors in this stage. The dependencies might include relationships such as co-occurrence or mutual exclusivity, as shown in the dependency column of Table~\ref{tab:noisefactor}. These usually depend on domain assumptions (e.g., obstructions and lens scratches often co-occur) and are determined in close consultation with domain experts. Once relevant noise factors and their dependencies are identified, we define discrete operational states for each factor. These may be binary (e.g., \emph{normal} vs.\ \emph{degraded}) or categorical (e.g., \emph{low}, \emph{normal}, \emph{high}), depending on the nature of the factor. For instance, lens scratches may be classified as: \emph{normal} (visually insignificant), \emph{low} (tolerable for AI perception), or \emph{high} (impairing object detection). These classifications are informed by expert input, empirical data, and the operational context specifications. Capturing factor dependencies and discretised states at this stage is essential to constrain combinatorial complexity in the situational coverage analysis conducted in Stage~2.

\subsection{Stage~2: Generating Situational Grid Coverage Table}
Situational grid coverage~\cite{Rob2015} offers a structured method for evaluating autonomous systems by mapping real-world conditions, such as lighting, obstacles, and human presence, into a multi-dimensional grid. Each cell in the grid represents a distinct subset of operating context, supporting systematic testing across a variety of environmental contexts to assess system safety and robustness. Unlike traditional coverage metrics, which focus on internal aspects such as code execution paths, situational coverage highlights external influences that affect system behaviour, particularly under complex or dynamic conditions~\cite{nawshin2023, proma2025scaloft}.

Stage~2 of our framework constructs situational coverage grids based on the noise factors, their discrete states, and known dependencies identified in Stage~1. The noise factor table (Table~\ref{tab:noisefactor}) is organised hierarchically across three levels: \emph{Type}, \emph{Noise Factor}, and \emph{State}. Situational coverage is applied at the most granular level by generating combinations of states within each type, while preserving logical consistency through dependency constraints. incorporating consistencies helps to prune infeasible combinations, thereby reducing final grid sizes. For the five noise factor types, (1) piece-to-piece variation, (2) change over time, (3) usage, (4) environment, and (5) system interactions, we obtain pruned grid sizes of $n_1 = 72$, $n_2 = 36$, $n_3 = 108$, $n_4 = 12$, and $n_5 = 24$, respectively. These values exclude invalid or contradictory state combinations, as reflected in Table~\ref{tab:noisefactor}.

As an illustrative example, Table~\ref{tab:coverageExample} presents the situational grid for the \textbf{Usage} type. This approach enables structured reasoning over a reduced but representative set of feasible situations, despite the theoretical infinitude of real-world scenarios. The resulting situational grid defines part of the operational contexts, associate with sensor degradation, in which robustness safety requirements will be derived, as described in Stage~3.

\subsection{Stage~3: Specifying Robustness Safety Requirement}\label{sec:robustness-requirements}
Assurance of Machine Learning (ML) for Autonomous Systems (AMLAS)~\cite{hawkins2021guidance} is a six-stage guidance framework for constructing a safety case for ML components. Stage two of AMLAS focuses on the development of ML safety requirements, which address both performance and robustness. Performance requirements define the expected functional behaviour of the ML model, while robustness requirements determine the range of operational contexts, including camera sensor degradation, under which these performance expectations must continue to hold.

We argue that performance and robustness aspects are inherently coupled and should not be treated separately. Instead, ML safety requirements should be formulated as performance requirements qualified by explicitly defined operational contexts. These operational contexts express the robustness dimension and are used to specify the range of conditions (e.g., sensor noise factors) within which the ML component must maintain safe performance. For instance, for a pedestrian detection task, we extend the performance safety requirement (RQ) in~\cite{gauerhof2020assuring} to include an operational constraint as follows:

\textbf{RQ:} When the ego vehicle is 50 metres from the crossing, the object detection component shall identify pedestrians that are on or near the crossing in their correct position under all conditions defined in \texttt{[PODs\#1]}.

In this formulation, \texttt{[POD\#1]} denotes a bounded operational context subset corresponding to the first row in the coverage table~\ref{tab:coverageExample}, where the performance requirement must hold. For example, this represents an operational context without camera degradation.

We propose that robustness requirements are best conceptualised as these POD specifications, which collectively define the valid operating envelope for a given set of performance requirements. The union of all PODs should span the complete intended operational context.

Expressing safety requirements (SR) in the form \emph{performance SRs \(+\) \texttt{[PODs]}}, where the performance requirements are explicitly qualified by the robustness context, enables targeted evaluation of ML models across well-defined slices of the Operational context. This approach enhances both the traceability and modularity of the resulting safety case, while supporting systematic robustness analysis.

\begin{table}[t!]
    \caption{Coverage grid for usage type factor}
    \label{tab:coverageExample}
\centering
    \centering
    \scriptsize
    \setlength{\tabcolsep}{2.5pt}
    \renewcommand{\arraystretch}{2}
    \begin{tabular}{|c|c|c|c|c|c|c|}
    \hline
    \textbf{ID} & \textbf{Misplacement} & \textbf{Impact} & \textbf{Contaminants} & \textbf{Obstructions} & \textbf{Lens Scratch} & \textbf{Dynamics} \\
    \hline
    1 & False & False & False & False & False & False \\
    2 & False & False & False & False & False & True \\
    3 & False & False & False & False & Low   & False \\
    %4 & False & False & False & False & Low   & True \\
    %5 & False & False & False & False & High  & False \\
    %6 & False & False & False & False & High  & True \\
    %7 & False & False & False & Low   & False & False \\
    %8 & False & False & False & Low   & False & True \\
    %9 & False & False & False & Low   & Low   & False \\
    ... & ... & ... & ... & ... & ... & ... \\
    108 & True & True & True & High & High & True \\
    \hline
    \end{tabular}
\end{table}

\section{Conclusion and Future Work}
This paper proposes a structured framework for deriving robustness safety requirements for AI-based perception systems affected by camera degradation in automotive settings. By combining noise factor identification with situation coverage analysis, the approach produces performance requirements tied to bounded operational contexts (\emph{SRs + [PODs]}), improving traceability and evaluation. Future work will address the scalability challenge by developing methods to prioritise the most critical PODs for efficient testing and validation.

\section*{Acknowledgment}
This work was supported by the Centre for Assuring Autonomy, a partnership between Lloyd’s Register Foundation and the University of York (https://www.york.ac.uk/assuring-autonomy/)

\bibliographystyle{ieeetr}  % Or another style like plain, unsrt, apalike, etc.
\bibliography{ref}          % Without the `.bib` extension

\vspace{12pt}
\end{document}